\documentclass[runningheads]{llncs}

\usepackage{eccvabbrv}
\usepackage{wrapfig}

\usepackage{graphicx}
\usepackage{booktabs}
\usepackage{siunitx}

\usepackage{customization}

\newcommand{\method}[1]{{\textbf{E-SSL\textsuperscript{3D}}}}

\usepackage[accsupp]{axessibility}
\usepackage{arydshln}
\makeatletter
\def\adl@drawiv#1#2#3{%
        \hskip.5\tabcolsep
        \xleaders#3{#2.5\@tempdimb #1{1}#2.5\@tempdimb}%
                #2\z@ plus1fil minus1fil\relax
                \hskip.5\tabcolsep}
\newcommand{\cdashlinelr}[1]{%
  \noalign{\vskip\aboverulesep
           \global\let\@dashdrawstore\adl@draw
           \global\let\adl@draw\adl@drawiv}
  \cdashline{#1}
  \noalign{\global\let\adl@draw\@dashdrawstore
           \vskip\belowrulesep}}
\makeatother
 \usepackage{multirow}

\usepackage{hyperref}

\usepackage{orcidlink}

\newcommand\blfootnote[1]{%
  \begin{NoHyper}
  \begingroup
  \renewcommand\thefootnote{}\footnote{#1}%
  \addtocounter{footnote}{-1}%
  \endgroup
  \end{NoHyper}
}

\begin{document}

\title{Equivariant Spatio-Temporal Self-Supervision for LiDAR Object Detection}

\titlerunning{\method~}

\author{Deepti Hegde\inst{1}$^\star$\and
Suhas Lohit\inst{2} \and 
Kuan-Chuan Peng\inst{2} \and \\
Michael J. Jones\inst{2} \and
Vishal M. Patel\inst{1} }

\authorrunning{Hegde et al.}

\institute{Johns Hopkins University
\\
 \and
Mitsubishi Electric Research Laboratories (MERL)\
}

\maketitle

\begin{abstract}
Popular representation learning methods encourage feature invariance under transformations applied at the input. However, in 3D perception tasks like object localization and segmentation, outputs are naturally equivariant to some transformations, such as rotation. Using pre-training loss functions that encourage \emph{equivariance} of features under certain transformations provides a strong self-supervision signal while also retaining information of geometric relationships between transformed feature representations. This can enable improved performance in downstream tasks that are equivariant to such transformations. In this paper, we propose a spatio-temporal equivariant learning framework by considering both spatial and temporal augmentations jointly. Our experiments show that the best performance arises with a pre-training approach that encourages equivariance to translation, scaling, and flip, rotation and scene flow. For spatial augmentations, we find that depending on the transformation, either a contrastive objective or an equivariance-by-classification objective yields best results. To leverage real-world object deformations and motion, we consider sequential LiDAR scene pairs and develop a novel 3D scene flow-based equivariance objective that leads to improved performance overall. We show our pre-training method for 3D object detection which outperforms existing equivariant and invariant approaches in many settings.
\end{abstract}

\blfootnote{$^\star$ This work was done when Deepti Hegde was an intern at MERL.}

\section{Introduction}

Depending on fully-supervised training paradigms can be limiting due to the expensive nature of manual annotations. Recent interest in autonomous navigation and the lowering cost of sensing hardware has enabled access to large amounts of LiDAR data \cite{mao2021one,waymo}, a crucial source of depth information useful for perception tasks. However, the annotation of outdoor LiDAR point clouds is challenging due to their irregularities and sparsity.  Self-supervised learning (SSL) enables the learning of generic visual representations of unlabelled data by completing tasks designed based on human intuition about what information can be inferred from its inherent properties, without the need for explicit supervision. The availability of large amounts of unlabelled LiDAR data thus makes SSL pre-training methods a natural choice for improving performance of perception tasks in limited label scenarios.

\begin{wrapfigure}{r}{0.5\textwidth}
    \centering
    \includegraphics[width=\linewidth]{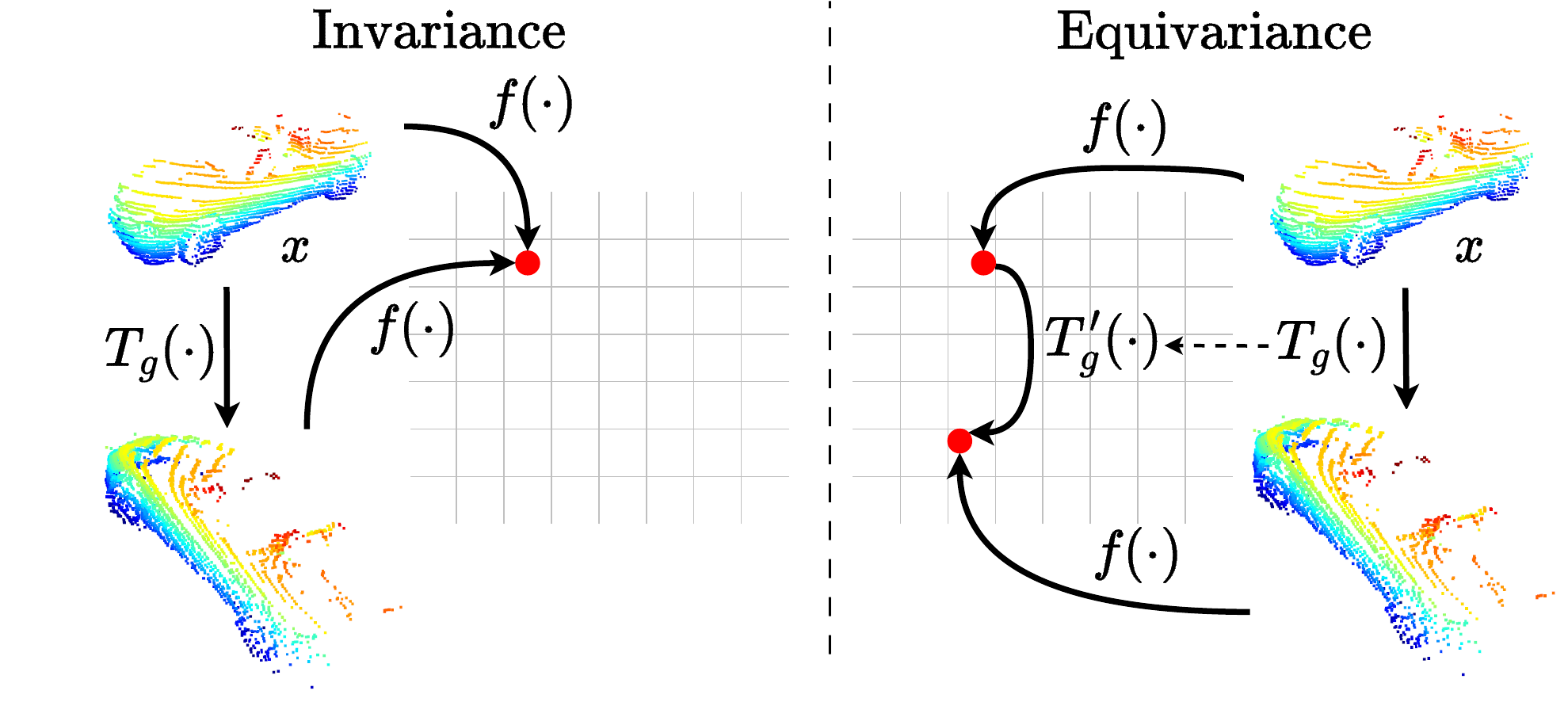}
    \caption{On the left, an illustration of invariance (as described in Eq. \ref{eq:inv}) vs. equivariance (as described in Eq. \ref{eq:equi}) on the right. Invariance of $f$ to group of transformations $G$ means that the output representation does not change with the applied input transformation, whereas equivariance of $f$ to $\mathcal{G}$  means the output representation changes by $T'_g$ for some applied input transformation $T_g$. In this  visualization, $T_g$ is 3D rotation. Figure based on \cite{devillers2022equimod}.}
    \label{fig:abs}
\end{wrapfigure}

The most popular pretext tasks of state-of-the-art (SOTA) representation learning approaches encourage feature invariance under different views, transformations, and across time by training the network under a contrastive learning objective \cite{he2020momentum,chen2020simple,xie2020self,yin2022proposalcontrast}. These frameworks empirically show which transformations should be selected to train the network to learn rich visual representations useful for downstream tasks \cite{chen2020simple,xiao2020should}.  Methods like STRL \cite{huang2021spatio} use a BYOL-like \cite{grill2020bootstrap} framework to encourage \textbf{invariance} of the features over time. However, in the tasks like object detection and semantic segmentation, the input orientation is important information that should be retained in the feature representation. That is, if the LiDAR scene is rotated, all the object bounding boxes should also rotate by the same amount. Encouraging invariance to rotations conflicts with this task. Therefore, in contrast to such methods, we propose an equivariant training objective, which encourages learning of the features to transform in an \textbf{equivariant} fashion over spatial augmentations as well as over time. In PointContrast \cite{xie2020pointcontrast}, networks are trained to be equivariant to rigid transformations such as rotation, scaling, and translation. In this work, we also show improved performance for rotation equivariance via equivariance-via-classification strategy.

We can define invariance and equivariance to group action more formally. Let the inputs be denoted by $x \in \cX$, $f(\cdot)$ be the encoder, and the output be $f(x)$. A group $G$ is a set along with a binary operation $\circ$ that respects closure ($\forall g, g' \in G, g \circ g' \in G$), associativity ($\forall g, g', g^{*} \in G, g \circ (g' \circ g^*) = (g \circ g') \circ g^*$), has a unique identity element $e \in G$ and an inverse $g^{-1}$ exists for each element $g \in G$ such that $g \circ g^{-1} = g^{-1} \circ g = e$. If $G$ is a group of transformations, for $g \in G$, let $T_g(x)$ and $T'_g(x)$ be the group action on $x$ and $f(x)$, respectively. 
The invariance of $f$ to $G$ means that the output representation does not change with the applied transformation, 
\begin{equation}
\label{eq:inv}
    \forall x,\ \forall g,\ f(T_g(x)) = f(x).
\end{equation}
By the equivariance of $f$ to $G$, we mean that
\begin{equation}
\label{eq:equi}
    \forall x,\ \forall g,\ f(T_g(x)) = T'_g(f(x)),
\end{equation}
where $T'_g$ is another transformation. Fig.~\ref{fig:abs} illustrates this difference. 

Dangovski \etal~\cite{dangovski2021equivariant} suggest that certain transformations that are discarded for the purpose of training for invariance can instead be leveraged to train for equivariance, and perform a series of experiments to validate this theory. However, the downstream tasks for image recognition includes only image classification, which is invariant to the transformations at the input. In this paper, we are interested in better understanding the effect of, and improving methods for equivariant pre-training for LiDAR point clouds for object detection that inherently has an equivariant component -- bounding box regression should be equivariant to geometric transformations applied at the input.

Data augmentation methods form an important component of self-supervised learning frameworks, and influence the nature of learned visual representations \cite{xiao2020should}. Existing augmentations for LiDAR scenes do not include realistic transformations that describe things such as ego-motion and geometric object deformations over time. For the visual perception of moving objects, relative motion becomes a useful property to localize objects. We expand the study of equivariant feature learning to more natural transformations that may be observed with time by considering temporal sequences of LiDAR frames. We model the point-level transformation over time as a scene flow matrix. Scene flow naturally captures local transformations of objects through their motion. 3D scene understanding tasks should be equivariant to these local transformations \cite{xiong2021self}. We thus include 3D scene flow as an additional transformation under which to train the network to be equivariant.

In this work, we present a study into \textbf{E}quivariance for \textbf{S}elf-\textbf{S}upervised \textbf{L}earning for \textbf{3D} perception tasks on LiDAR point clouds and propose the framework \method~. We consider LiDAR scenes applied with a series of spatial and temporal augmentations to train a 3D feature encoder under a joint equivariant contrastive learning and flow equivariance objective. To encourage spatial equivariance, transformed views of a scene are contrasted at the point level as well as passed to a classification head to predict the applied geometric transformation. To encourage temporal equivariance, the network is trained to minimize the distance between sequential pairs of LiDAR frames in the voxel feature space, where the feature map of the first frame is warped to the second frame. 

We conduct extensive experiments on 3D object detection in low-data training scenarios. Our contributions are as follows:
\begin{enumerate}
    \item We propose \method~, a self-supervised pre-training method for LiDAR scenes that trains a network to learn spatio-temporal equivariance through a joint loss objective. We are the first to leverage 3D scene flow to encourage equivariance to temporal augmentations in LiDAR scenes.
    
    \item We show that our pre-training strategy is effective in improving the performance of downstream tasks, particularly in low-data scenarios. We show that an object detection network pre-trained using \method~ and fine-tuned using just 20\% of data can achieve comparable performance to a network trained from scratch on 100\% data. 
    \item We show improved performance for rotation equivariance over standard contrastive approaches through an equivariance-via-classification strategy. 
\end{enumerate}

\section{Related works}

\subsection{Equivariant self-supervision}

In recent years, there has been a growing interest in exploring the role of equivariance in learning visual representations in a self-supervised manner \cite{xie2020self,dangovski2021equivariant,gupta2023learning,garrido2023self,devillers2022equimod,bhardwaj2023steerable}. Dangovski \etal generalize the standard contrastive SSL framework to a equivariant SSL framework and improve the existing performance of purely invariant SSL methods on the tasks such as image classification and regression problems in photonics. In this work, we follow the intuition of probing for complementary augmentations, but apply equivariant SSL to more complex downstream tasks that are inherently equivariant, such as object detection. In \cite{garrido2023self}, Garrido \etal propose a benchmark for evaluating equivariance on both inherently equivariant tasks such as 3D rotation prediction, as well as invariant ones such as classification, and present a method of splitting representations into invariant and equivariant parts. CARE \cite{gupta2023learning}  learns to translate augmentations such as cropping into linear transformations in a spherical feature space. The above approaches evaluate their frameworks on the tasks that generally benefit from invariance, such as classification or on the benchmarks specifically designed to evaluate learned equivariance. However, little investigation has been made into the pretext tasks that encourage equivariance for improving downstream tasks for 3D scene understanding. Xiong \etal propose FlowE \cite{xiong2021self}, an SSL framework for image segmentation and object detection that is a variation of BYOL \cite{grill2020bootstrap}, where they introduce a flow equivariance objective by applying the flow transformation to a reference video frame, thus covering natural deformations. However, this approach is only applied to image video sequences. To the best of our knowledge, we are the first to explore the role of equivariance to both spatial and temporal, synthetic and natural  transformations for 3D scene understanding with LiDAR. 
 
\subsection{Self-supervision for point cloud scenes}
Self-supervised pre-training shows promise for learning useful representations from unlabeled point cloud scenes, both indoors and outdoors. Xie \etal propose PointContrast \cite{xie2020pointcontrast}, which point-level contrastive training across partial transformed views of indoor scenes. This objective encourages equivariance to rigid transformations. In this work, we explore more effective ways of learning equivariant features under the more natural augmentation of 3D scene flow, as well as including an equivariance-via-classification strategy.  Several follow-up works such as SegContrast \cite{nunes2022segcontrast} and  DepthContrast \cite{zhang2021self} contrast point-level and segment-level features of transformed point clouds. The methods specifically tailored to outdoor LiDAR point clouds leverage contrastive learning, occupancy prediction, and point cloud completion to learn meaningful representations. TARL \cite{nunes2023temporal} learns temporally consistent feature representations by associating objects across time and maximizing their feature similarity. This approach depends on indexing and spatial clustering to compute object correspondence, which can be a limiting factor when considering LiDAR sequences captures with a high frame rate. ALSO \cite{boulch2023also} employs surface reconstruction as an auxiliary task to improve downstream performance in object detection and semantic segmentation. This pre-training strategy is specific to the downstream task and network, and thus lacks general applicability. Our pre-training framework is unified for all downstream networks that share a 3D feature backbone. ProposalContrast \cite{yin2022proposalcontrast} considers region-level features obtained through a spatial clustering approach to enforce feature similarities between transformed objects. However, the success of their method depends on the quality of unsupervised region clustering, which becomes difficult for sparse LiDAR scenes.

The above approaches perform self-supervised learning using convolutional feature backbones. Recent approaches propose the pretext tasks for representation learning to improve the performance of transformer-based 3D object detection networks. Yang \etal propose GD-MAE \cite{yang2023gd}, a generative approach based on the masked autoencoder (MAE) \cite{he2021masked}, which uses hierarchical fusion to infer information from masked voxels. MV-JAR \cite{xu2023mv} employs both masked reconstruction and voxel position estimation in the form of a classification objective to perform self-supervised learning on LiDAR scenes. These approaches are effective as pre-training strategies for transformer-based detection networks such as SST \cite{fan2022embracing}, however cannot be applied to sparse convolutional backbones. We provide a more general solution for representation learning for LiDAR scenes, and focus on feature extractor backbones that are sparse convolution based, as the same backbone may then be used for a large number of detection \cite{shi2020pv,second,shi2019pointrcnn,lang2019pointpillars,deng2021voxel,shi2021parta2} and segmentation \cite{choy20194d,tang2020searching,zhu2021cylindrical} networks.

\subsection{3D object detection}

The neural networks that perform 3D object detection on LiDAR scenes process point clouds as points \cite{shi2019pointrcnn,yang20203dssd}, as a set of 3D grids known as voxels \cite{second,lang2019pointpillars,deng2021voxel}, or a combination of the two \cite{shi2020pv,shi2023pv}. Single stage networks \cite{second,yang20203dssd} directly estimate bounding box dimensions and predict category labels whereas two stage networks include an additional bounding box refinement head \cite{shi2019pointrcnn,deng2021voxel}. The popular detection network SECOND \cite{second} is a single stage detector consisting of a 3D sparse convolution backbone and a 2D convolutional layer following a Bird's-Eye View (BEV) compression step. The two-stage detector VoxelRCNN \cite{deng2021voxel} shares a similar architecture but has an additional region refinement head and proposes a novel region pooling approach. Despite differences in architectures many sparse-convolutional 3D detectors share a common 3D feature extractor, which is advantageous for pre-training. Recently, attempts have been made to move away from the sparse convolutional approaches that operate on voxels and move towards transformer-based backbones \cite{fan2022embracing}, but these methods have high memory requirements. In this work, we focus on the detectors with convolutional backbones due to the broader applicability to downstream tasks. A single pre-trained 3D backbone is applicable to multiple detectors.

\begin{figure*}[!h]
    \centering
    \includegraphics[width=\linewidth]{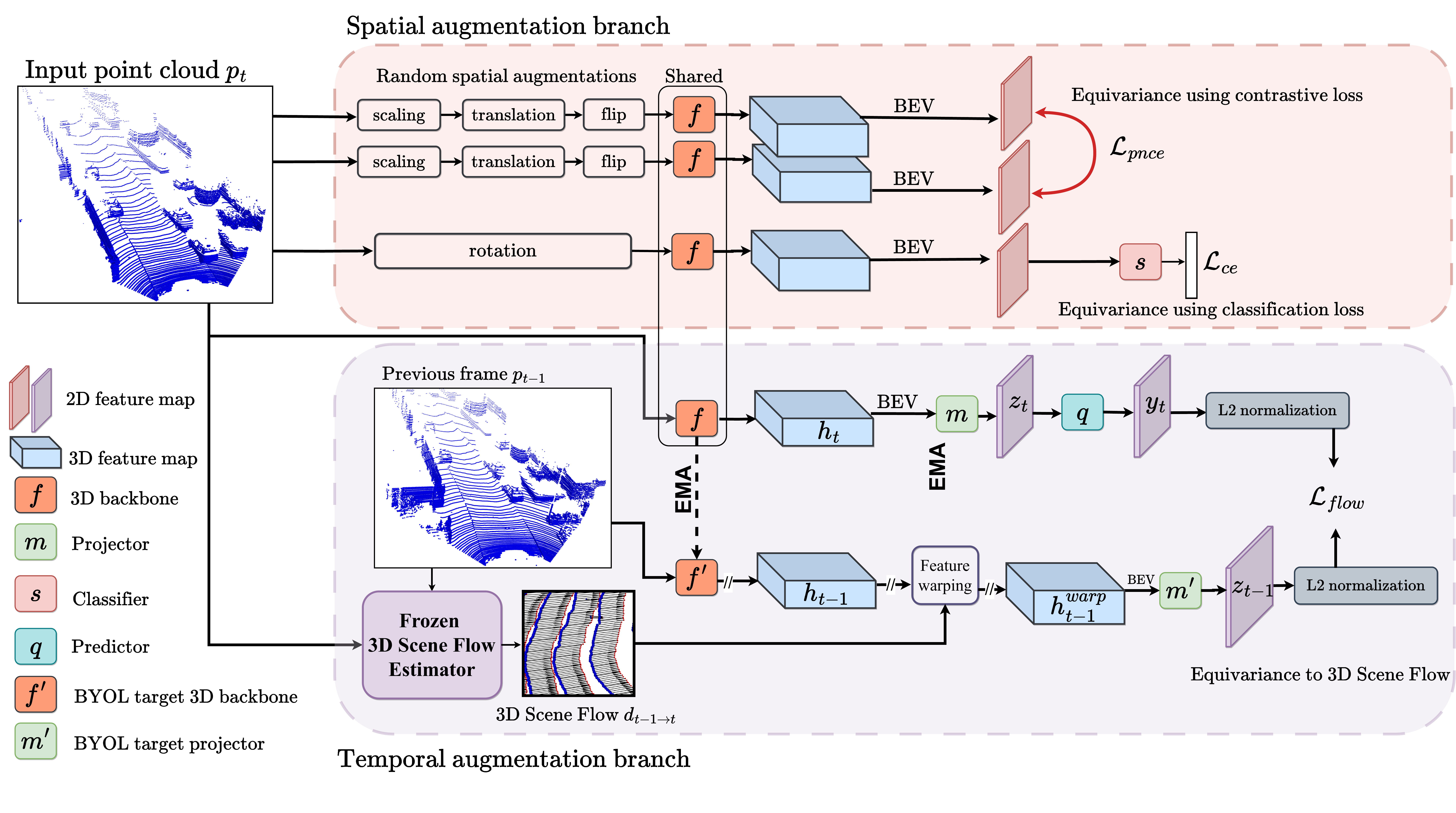}
    \caption{An overview of \method. A LiDAR point cloud undergoes spatial and temporal augmentations before being input to the network that consists of a 3D feature extraction backbone $f$, a projector network $m$, a predictor network $q$, and a classifier $s$. ($f$,\ $m$,\ $q$) form the online branch and the copies ($f'$,\ $m'$) form the target network that is only updated through an exponential moving average (EMA) of the weights of the online network.}
    \label{fig:arch}
\end{figure*}

\section{Proposed method: \method~}

Our goal is to train a network to be equivariant to certain spatial and temporal transformations in order to  learn meaningful geometric representations that aid the network in performing dense scene prediction tasks. \method~ trains a feature encoder to learn dense voxel-level features that reflect the natural deformations that arise from object motion and that are equivariant to rigid transformations as well as naturally occurring temporal augmentations. This is done by a joint training procedure that consists of learning equivariant features for both spatial augmentations and temporal changes via computing 3D scene flow. Our overall framework is summarized in Fig.~\ref{fig:arch}.

Consider an input point cloud $p$ from a sequence of LiDAR scenes. The network that is being trained using self-supervision can be divided into four parts -- (a) a 3D feature encoder $f$ that is a point-based or a voxel-based neural network that maps the input point cloud to a set of 3D features which are then mapped into 2D BEV features, (b) a projector $m$ that maps the BEV features into a lower feature dimension, (c) a predictor $q$ that matches the output of the projector to a target feature map, and (d) an augmentation classifier $s$. 

\subsection{Equivariance to global rigid spatial augmentations}
We focus on common global rigid geometric transformations that are invertible -- rotations, translations, scaling, and flips. We train the network to be equivariant to a group of spatial transformations in two ways, through a contrastive training objective and through an augmentation prediction objective.

Let the set of rigid transformations be $\mathcal{A}$. During training, for each iteration and each input point cloud, two instances from $\mathcal{A}$ are sampled randomly and applied to the input point cloud. We apply the contrastive loss PointInfoNCE \cite{xie2020pointcontrast} at the point level, which encourages invariance to the applied rigid transformations through a point-level contrastive objective between matched points in a scene. Considering the two augmented views, Let $\cP^+$ be the set of all positive matched points from both views. For a matched pair $(i,j) \in \cP^+$:
\begin{equation}
    \cL_{pnce} = -\sum_{(i,j)\in\cP_+} \log\frac{\exp((\textbf{x}_i \cdot  \textbf{x}_j)/\tau )}{\sum _{(\cdot ,k)\in \cP_+}\exp((\textbf{x}_i \cdot  \textbf{x}_k)/\tau)},
\end{equation}
where $(\textbf{x}_i,\textbf{x}_j)$ are the point feature vectors for each scene and $\tau$ is the scaling parameter. 

Inspired by \cite{dangovski2021equivariant}, we consider a uniform discrete subset of rigid transformations to further encourage equivariance through a classification objective. For example, for planar rotations, we consider 10 randomly sampled rotations. The features extracted for these augmentations are passed through an additional classifier layer $s$ that is trained to predict the transformation applied at the input. The parameters of $f,\  g$, and $s$ are trained by minimizing the cross-entropy loss $\cL_{ce}$ between the input transformation and the predicted transformation. Thus, the network is trained to retain the information of the transformation that was applied at the input, \ie, the features are equivariant to the input transformation. Note that the cross-entropy loss is applied in addition to the contrastive loss.

\subsection{Equivariance to temporal changes via 3D scene flow estimation and feature warping}

While spatial augmentations help in improving the final downstream performance, the augmentations are not realistic, in general. In addition to spatial augmentations, we exploit real frames that are captured sequentially to provide additional self-supervision. The key insight for temporal self-supervision is that the features of the network should evolve in an equivariant manner to how the points move in the real world. We term this 3D scene flow equivariance, where the 3D scene flow is the vector field that describes the motion of points in a point cloud frame at some time instance to locations in a different time instance. The 3D scene flow is therefore a locally varying transformation computed between two real frames and the features learned by the network are trained to respect the 3D scene flow constraint. 

\noindent\textbf{Estimating 3D scene flow.}
Given the point cloud $p$, we consider its natural temporal augmentation by taking the previous frame in the sequence. For the purpose of illustrating scene flow, let the previous frame be denoted at $p_{t-1}$ and $p_t$ be the current frame. We model the temporal transformation between scenes as 3D scene flow, represented as per-point displacement, $d_{t-1 \rightarrow t} \in \RR^3$ for each point in the point clouds.  We denote the forward time transformation operation as $\cF_{t-1 \rightarrow t}$. Thus, we can represent the relationship between $p_{t}$ and $p_{t-1}$ as: 
\begin{equation}
    p_{t} = \cF_{t-1 \rightarrow t}(p_{t-1}).
\end{equation}
$d_{t-1 \rightarrow t}$ is estimated from a frozen 3D scene flow estimation network based on PV-RAFT \cite{wei2021pv} trained on synthetic data and adapted to real-world LiDAR data in an unsupervised manner through a student-teacher framework \cite{jin2022deformation}.

\noindent\textbf{Learning flow equivariance.}
We build the flow equivariance component on BYOL \cite{grill2020bootstrap}. Under this SSL framework, a set of two networks -- online and target -- are trained to minimize the distance between predicted feature maps. Given an augmented view, the online network is trained to predict the representation of the scene from the target network under a different augmentation. We choose this framework as an alternative to the contrastive learning approach since in the case of natural augmentations such as object motion across time, there is no useful discriminative feature to be learned by considering ``negative" samples. This strategy for training for flow equivariance has seen success for images \cite{xiong2021self}. The information of the architectures of the online and target networks is in Sec.~\ref{sec:pre}. An illustration of the online-target architecture can be seen in Fig.~\ref{fig:arch}. 

The 3D feature backbone $f$, the projector $m$, and the feature predictor  $q$ are considered as the online network, while the copies $f'$ and $m'$ are considered as the target network. Note that $f$ is shared with the spatial augmentations branch. The role of $m$ is to create more general representations for ease of adaptation of $f$ to downstream tasks. The feature predictor matches the representation from the online network to that of the target network. The online network is updated with the standard training process while the weights of the target network are updated only through an exponential moving average (EMA) of the online network. This prevents representation collapse during the prediction step. Let the parameters of the online network as a whole be denoted as $\theta$, and that of the target network be $\xi$. The EMA weight update step can be written as: 
\begin{equation}
    \xi \ \leftarrow \ \gamma\xi + (1 - \gamma)\theta,
\end{equation}
where $\gamma \in [0,1]$ is the target decay rate. $\gamma$ is updated at every training step using the formula: 
\begin{equation}
    \gamma \triangleq 1 - (1 - \gamma_{\text{base}}) \cdot (\cos(\pi k/K) + 1)/2,
\end{equation}
where $k$ is the current training step and $K$ is the maximum number of training steps, with $\gamma_{\text{base}}$ being the initial target decay rate. 

The point cloud scene pair $(p_{t-1},\ p_{t})$ is fed to the target network and online network, respectively. The 3D voxel feature representations $h_{t-1}$, $h_{t}$ are obtained by 
\begin{equation}
    h_{t} = f(p_{t}),\
    h_{t-1} = f'(p_{t-1}).
\end{equation}
$h_{t-1}$ is warped to the future feature frame $h^{warp}_{t-1}$ using the 3D scene flow estimate in a process detailed in the next section. After BEV compression, the features ($h^{warp}_{t-1},\ h_{t}$) are passed to the projectors $m'$ and $m$ respectively to give ($z_{t-1},\ z_{t}$). 
The projected feature map $z_{t}$ is the input to feature predictor $q$ to give $y_{t}$ which is matched to the output of the target network $z_{t-1}$.

\noindent\textbf{Warping with 3D scene flow.}
 At the input, the point cloud $p_{t-1}$ can be transformed to the next scene, point cloud $p_{t}$, by simply adding the per-point displacement from the scene-flow matrix, assuming index correspondence is maintained, which holds for the datasets used \cite{nunes2023temporal}. Let this warped estimate at the input be $p^{warp}_{t}$. Performing the same transformation in the feature space is not as straightforward, as we deal with the 3D sparse spatial feature maps $(h_{t-1},\ h_{t})$ instead of sets of points. The 3D feature maps are represented as sparse tensors, which consist of voxel features and their corresponding 3D coordinates denoting their position in the feature volume. From the flow estimate $p_{1 \rightarrow 2}$, we estimate the voxel coordinates of the points of the estimated future frame by applying the standard input voxelization process on $p^{warp}_{t}$. These estimated future voxel coordinates are then used to sample voxel features from $h_{t-1}$ to get the warped feature estimate $h^{warp}_{t-1}$.

\noindent\textbf{Loss function.}
The loss objective is the minimization of the $L2$ distance between the normalized $z_{t-1}$ and $y_{t}$ averaged over the spatial dimensions, defined as:
\begin{equation}
    \cL_{flow} = \frac{1}{HW}\left \| \hat{z}_{t-1} - \hat{y}_{t} \right \|_{2}^2,
\end{equation}
where $HW$ is the spatial dimension and ($\hat{z}_{t-1},\ \hat{y}_{t}$) are the normalized feature maps.

The final loss function used to train the network is the sum of the contrastive loss and $L2$ distance loss, both of which are brought to similar scales through loss coefficients $\lambda_{pnce}$, $\lambda_{ce}$ and $\lambda_{flow}$ (\ie, we set the loss coefficients such that all the loss terms have comparable ranges). The final loss is written as:
\begin{equation}
    \cL = \lambda_{pnce}\cL_{pnce} + \lambda_{ce}\cL_{ce} + \lambda_{flow}\cL_{flow},
\end{equation}

\section{Experiments}

\subsection{Pre-training}
\label{sec:pre}

\noindent\textbf{Datasets.} 
We use the KITTI-360 \cite{Liao2022PAMI} and SemanticKITTI \cite{behley2019iccv} datasets for the purpose of pre-training. KITTI-360 consists of 100k LiDAR scenes from 11 sequences captured in urban roads. SemanticKITTI contains 22 sequences and roughly 48k scans. For the purpose of pre-training, we remove the validation sequences. To mitigate the distribution gap between the pre-training and fine-tuning datasets, we consider the front field-of-view (FFOV) scenes during pre-training.

\noindent\textbf{Augmentation.}
As established, we choose the spatial augmentations that are invertible, namely global rotation about the vertical axis with an angle in the range $(\frac{-\pi}{2},\ \frac{\pi}{2})$, global translation in the $(x,\ y,\ z)$ axes with the displacement range ($0m,\ 0.2m$), global scaling with magnitudes falling in the range ($0.95,\ 1.05$), and random vertical flip with a probability of $50\%$. We choose to train the network to be equivariant to rotations, based on the experiments in Sec.~\ref{sec:prelim}.  For ease of prediction, we sample from a discrete set of 10 rotation angles. For the temporal augmentation, we sample the previous frame from the sequence of LiDAR frames. KITTI-360 consists of LiDAR sequences that capture around 1.2 frames for every meter, with a 10 meter overlap between consecutive frames. SemanticKITTI consists of sequences that capture 10 frames per second.

\noindent\textbf{Network architecture.}
In this section, we detail the network architectures of each module of the the proposed framework. 

\noindent\ul{3D feature encoder ($f$)}:
We perform pre-training on the sparse convolutional feature backbone SparseVoxel backbone popular among recent 3D object detection networks \cite{second,deng2021voxel,lang2019pointpillars}. 

\noindent\ul{Projector ($m$)}:
We perform feature projection on the 2D BEV compressions of 3D volumetric feature maps. The BEV feature maps are obtained by max-pooling the densified sparse tensor along the height dimension. The feature projector is a 3-layer 2D convolutional network with batch normalization and ReLU layers. It maintains the spatial dimensions of the BEV feature map while reducing the channel dimension from $256$ to $128$. 

\noindent\ul{Classifier ($s$)}:
The classification branch of the network predicts the applied $n$-fold transformation. This is a simple 3-layer fully connected network with 2 batch normalization layers. In practice, we choose $n=10$. 

\noindent \ul{Predictor ($q$)}:
The predictor network of the online branch consists of a single 1$\times$1 convolutional layer to maintain the spatial dimensions.

\noindent\textbf{Implementation details.}
We use the AdamW optimizer with a cyclic learning rate schedule, with the maximum learning rate $10^{-4}$, a weight decay of $0.01$, and momentum $0.9$. The network is trained for 80 epochs with a batch size of 56 split over 8 NVIDIA A6000 GPUs. We use the codebase OpenPCDet \cite{openpcdet2020} for the implementation of the 3D encoder and BEV projection modules. We follow \cite{grill2020bootstrap} for the hyperparameters and EMA update rules of the online-target networks, with an initial $\gamma_{base} = 0.999$. We use the implementation of PointContrast for LiDAR scenes from the 3DTrans codebase \cite{yan2023spot}. Here, point features are sampled from the multi-scale 3D features and the compressed BEV feature map. We modify the computation of the point features to include the output after projection. We samples 2048 points for both the positive and negative samples. The values for the loss coefficients are $\lambda_{pnce} = 0.01$, $\lambda_{flow}=300$, $\lambda_{ce}=1$.

\subsection{Object detection experimental details}
We demonstrate the effectiveness of our pre-training strategy for 3D object detection using the two detectors SECOND \cite{second} and VoxelRCNN \cite{deng2021voxel}. We fine-tune these object detectors on the KITTI object detection dataset \cite{KITTI} under the standard training and validation splits, and perform evaluation using the official metrics. We perform fine-tuning in 4 data availability scenarios, 5\%, 20\%, 50\%, and 100\% of data. To account for class bias, we perform subset sampling thrice for each split and report the average performance.

\noindent\textbf{Network architectures.}
SECOND \cite{second} is a single stage detector that consists of a sparse convolution 3D encoder, a BEV encoder, and a region proposal network. VoxelRCNN is a two stage network that shares a similar 3D backbone and 2D encoder but includes an additional proposal refinement head.

\noindent\textbf{Datasets and metrics.}
We fine-tune the object detection networks on the KITTI object detection dataset \cite{KITTI}, which consists of 3712 training samples and 3769 validation samples. We perform evaluation under the standard protocol detailed in \cite{KITTI} on three difficulty categories and report performance on the ``Car", ``Pedestrian", and ``Cyclist" categories as well as the mean average precision. Precision is calculated under 40 recall positions, as is followed in \cite{boulch2023also}. We use the standard division of objects into their respective difficulties based on their truncation, occlusion, and distance from the camera.

\begin{table*}[h]

\centering
\resizebox{\textwidth}{!}{%
\begin{tabular}{cccccccccccc}
\toprule
\multirow{3.5}{*}{Split} & \multirow{3.8}{*}{Method} & \multicolumn{9}{c}{average precision (AP) (\%)}                                             & \multirow{3.8}{*}{mAP (\%)} \\ \cmidrule{3-11}
                       &                                              & \multicolumn{3}{c}{Car} & \multicolumn{3}{c}{Pedestrian} & \multicolumn{3}{c}{Cyclist} &                      \\ \cmidrule(l){3-5} \cmidrule(l){6-8} \cmidrule(l){9-11}
                       &                                              & easy   & moderate   & hard  & easy     & moderate     & hard     & easy     & moderate    & hard    &                      \\ \midrule
\multirow{5}{*}{5\%}   & No pre-training                              & 86.40  & 73.01  & 67.79 & \ul{42.76}    & \bf{38.26}    & \textbf{34.42}    & 64.13    & 45.44   & 42.70   & 54.99                \\
                       & PointContrast                            & \ul{86.93}  & 73.48  & 69.30 & 41.11    & 37.67    & 34.34    & 64.31    & 47.26   & 44.21   & 55.40                \\
                       & STRL                                     & 86.61  & 72.90  & 68.37 & 39.24    & 35.46    & 31.80    & 59.90    & 42.52   & 39.66   & 52.94                \\
                       & ALSO                                     & 86.80  & \textbf{75.56}  & \textbf{72.88} & 41.92    & 36.90    & 34.02    & \textbf{74.72}    & \textbf{58.99}   & \textbf{55.42}   & \textbf{59.69}                \\
                       & \method~                                  & \textbf{87.53}  & \ul{74.96}  & \ul{70.47} & \textbf{43.30}    & \ul{37.71}    & \ul{34.40}    & \ul{74.02}    & \ul{54.13}   & \ul{50.91}   & \ul{58.60}                \\ \midrule
\multirow{5}{*}{20\%}  & No pre-training                              & 87.64  & 77.12  & 73.13 & 50.36    & 45.70    & 41.29    & 75.26    & 54.99   & 51.24   & 61.86                \\
                       & PointContrast                            & 88.03  & 77.49  & 73.15 & 50.45    & 45.97    & 41.21    & 74.07    & 54.48   & 50.65   & 61.72                \\
                       & STRL                                     & 87.97  & 77.50  & 73.25 & \ul{51.68}    & \ul{46.82}    & 42.00    & 72.49    & 53.89   & 50.08   & 61.74                \\
                       & ALSO                                     &  \ul{88.73} & \textbf{79.44}  & \textbf{76.32} &  \textbf{52.46}   &  \textbf{47.71}   &   \textbf{44.28}  &   \textbf{81.74}  &  \textbf{65.50}  &  \textbf{61.31} &   \textbf{66.39}              \\
                       & \method~                                 &  \textbf{89.83}   & \ul{78.70}  & \ul{75.93} &   51.65  &  46.10   &  \ul{42.40}   & \ul{80.83}    &  \ul{62.92}  & \ul{59.13}   & \ul{65.28} \\ \midrule 
\multirow{5}{*}{100\%} & No pre-training                              & 90.05  & 81.03  & 78.09 & 53.97    & 49.19    & 44.06    & 80.59    & 63.54   & 59.64   & 66.69                \\
                       & PointContrast                            & 88.40  & 80.50  & 76.44 & 53.02    & 48.34    & 44.01    & 80.05    & 62.32   & 58.61   & 65.74                \\
                       & STRL                                     & \ul{90.37}  & 81.11  & 78.21 & \bf{59.55}    & \bf{53.61}    & \bf{48.08}    & 78.72    & 62.47   & 57.92   & 67.78                \\
                       & ALSO                                     & \textbf{90.44}  & \textbf{81.66}  & \textbf{78.83} & 56.41    & 51.91    & 47.53    & \ul{84.49}    & \ul{67.65}   & \ul{63.53}   & \ul{69.18}                \\
                       & \method~                                  &   90.14     &   \ul{81.64}     &   \ul{78.61}    &    \ul{57.53}      &     \ul{53.05}     &     \ul{47.72}       &     \textbf{85.40}   &     \textbf{69.54}     &  \textbf{64.69}     &    \textbf{69.81}                  \\  \bottomrule
\end{tabular}
} 
\caption{3D object detection with SECOND \cite{second} pre-trained on KITTI-360 \cite{Liao2022PAMI} and fine-tuned on KITTI \cite{KITTI} under different data splits. Each result is an average over 3 fixed subsets of the dataset. We report 3D average precision for 3 categories as well as the mean average precision over 40 recall positions. The best and second best performance is marked in \bd{bold} and \ul{underline}, respectively.}

\label{tab:second}
\end{table*}

\begin{table*}[h]
\centering
\resizebox{\textwidth}{!}{%
\begin{tabular}{ccccccccccccc}
\toprule 
\multirow{3.5}{*}{Split} & \multirow{3.8}{*}{Method}  & \multicolumn{9}{c}{average precision (AP) (\%)}                                             & \multirow{3.8}{*}{mAP (\%)} \\ \cmidrule{3-11}
                       &                                             & \multicolumn{3}{c}{Car} & \multicolumn{3}{c}{Pedestrian} & \multicolumn{3}{c}{Cyclist} &                      \\ \cmidrule(l){3-5} \cmidrule(l){6-8} \cmidrule(l){9-11}
                       &                                               & easy   & moderate   & hard  & easy     & moderate     & hard     & easy    & moderate    & hard    &                      \\ \midrule
\multirow{5}{*}{5\%}   & No pre-training                              & 88.89  & \ul{79.21}  & 75.55 & \textbf{57.50}    & \textbf{49.84}    & \ul{44.27}    & 78.92   & 59.73   & 55.97   & 65.54                \\
                       & PointContrast                            & \textbf{89.94}  & \ul{79.21}  & \bf{76.12} & 56.13    & 48.13    & 43.01    & 77.98   & 58.92   & 55.20   & 64.96                \\
                       & STRL                                     & 89.30  & 78.92  & \ul{75.94} & 55.68    & 48.13    & 42.73    & 73.98   & 56.85   & 53.26   & 63.87                \\
                       & ALSO                                     & \ul{89.74}  & \textbf{79.37}  & 75.91 & \ul{56.33}   & \ul{49.79}    & \textbf{44.77}    & \ul{82.84}   & \ul{64.09}   & \ul{60.16}   & \textbf{67.00}                \\
                       & \method~                                  &  88.79      &    78.93    &    75.41   &   56.02       &  48.55        &    43.19      &  \textbf{82.85}       &   \textbf{64.40}      &      \textbf{60.53}   &  \ul{66.52}                     \\  \midrule
\multirow{5}{*}{20\%}  & No pre-training                              & 91.99  & 82.10  & 79.40 & 56.09    & 49.29    & 44.26    & 85.24   & 67.55   & 63.13   & 68.78                \\
                       & PointContrast                            & 92.23  & 82.25  & 79.57 & 57.33    & 50.74    & 45.43    & 84.16   & 66.74   & 62.28   & 68.97                \\
                       & STRL                                     & 91.97  & 82.07  & 79.41 & 57.40    & 50.85    & 45.38    & \ul{86.36}   & 68.64   & 64.23   & 69.59                \\
                       & ALSO                                     & \ul{92.46}  & \bf{82.44}  & \ul{79.77} & \ul{60.57}    & \ul{53.21}    & \ul{48.61}    & 86.22   & \ul{69.88}   & \ul{65.40}  & \ul{70.95}                \\
                       & \method~                                  & \textbf{92.67}  & \ul{82.42}  & \textbf{79.89} & \textbf{60.72}    & \textbf{53.94}    & \textbf{49.19}    & \textbf{88.04}  & \textbf{71.40}   & \textbf{66.36}   & \textbf{71.63}                \\                 \midrule
\multirow{5}{*}{100\%} & No pre-training                              & \ul{92.45}  & \textbf{83.00}  & \ul{80.20} & \textbf{62.41}    & \textbf{55.89}    & \textbf{50.31}    & 88.40   & 68.81   & 64.42   & 71.77                \\
                       & PointContrast                            & 91.73  & 82.41  & 79.89 & 59.82    & \ul{54.14}    & 48.54    & 87.28   & 69.15   & 63.54   & 70.72                \\
                       & STRL                                     & 92.27  & 82.54  & 79.99 & \ul{61.38}    & 54.01    & 48.31    & 86.95   & 67.64   & 63.31   & 70.71                \\
                       & ALSO                                     & \textbf{92.57}  & \ul{82.88}  & \textbf{80.24} & 60.10    & 52.12    & 46.76    & \ul{90.71}   & \textbf{73.94}   & \ul{69.21}   & \ul{72.06}                \\
                       & \method~                                  &   92.08     &    82.73    &  80.18     &  61.00        &     53.82     &   \ul{48.58}       &    \textbf{91.15}    &   \ul{72.68}      &    \textbf{69.32}     &      \textbf{72.41}                \\  \bottomrule
\end{tabular}
} 
\caption{3D object detection with VoxelRCNN \cite{deng2021voxel} pre-trained on KITTI-360 \cite{Liao2022PAMI} and fine-tuned on KITTI \cite{KITTI} under different data splits. Each result is an average over 3 fixed subsets of the dataset. We report 3D average precision for 3 categories as well as the mean average precision over 40 recall positions. The best and second best performance is marked in \bd{bold} and \ul{underline}, respectively. }
\label{tab:voxelrcnn}
\end{table*}

\noindent\textbf{Implementation details.}
We use the AdamW optimizer with a cyclic learning rate schedule, with the maximum learning rate $3\times 10^{-3}$ for SECOND and VoxelRCNN. We use a weight decay of $0.01$, and momentum $0.9$. We use the codebase OpenPCDet \cite{openpcdet2020} for the implementation of the detection networks. Each network is fine-tuned for 80 epochs with a batch size of 8 over 2 NVIDIA A6000 GPUs. The temperature parameter $\tau$ in Eq. 3 in the main paper is $\tau = 1$. The value of the target decay rate for the exponential moving average described in Eq. 6 of the main paper is  $\gamma_{base}=0.9996$.

\noindent\textbf{Comparative methods.}
We compare the performance of \method~ with the recent SOTA SSL methods for LiDAR scenes:
\begin{itemize}
    \item \textbf{PointContrast} \cite{xie2020pointcontrast} is a representation learning method for point cloud scenes that encourages point-level equivariance to different transformed views. We perform spatial augmentations to create view pairs and implement the adapted version for LiDAR scenes that samples point-level features from multi-scale 3D and BEV features. We sample 2048 points. 
    \item \textbf{STRL} \cite{huang2021spatio} encourages feature invariance across synthetically created temporal sequences of point cloud scenes by minimizing the $L2$ distance between samples passed through a BYOL-like online-target network. This becomes an invariant counterpart to our temporal equivariance component, and we re-implement this method for LiDAR scenes by training the online-target networks with sequential scene pairs. 
    \item \textbf{ALSO} \cite{boulch2023also} is a recent SSL approach for LiDAR scenes that uses occupancy prediction as a pretext task. We use the model pre-trained on KITTI-360 for the SECOND detector, and reproduce the fine-tuning result to the best of our ability. We note that our reproduction is slightly lower than reported.  We note that this is a generative representation learning approach that differs fundamentally from our discriminative one.
    
\end{itemize}

\begin{table*}[]
\centering
\resizebox{\linewidth}{!}{%
\begin{tabular}{cccccccccccc}
\toprule
\multirow{3.8}{*}{\shortstack{Spatial\\equivariance}} & \multirow{3.8}{*}{\shortstack{Temporal\\equivariace}} & \multicolumn{9}{c}{average precision (AP) (\%)}                                                                                                                                                                                                                 & \multirow{3.8}{*}{mAP(\%)}      \\ \cmidrule(l){3-11}
                                      &                                        & \multicolumn{3}{c}{Car}                                                           & \multicolumn{3}{c}{Pedestrian}                                                    & \multicolumn{3}{c}{Cyclist}                                                      &                           \\ \cmidrule(l){3-5} \cmidrule(l){6-8} \cmidrule(l){9-11}
                                      &                                        & \multicolumn{1}{c}{easy} & \multicolumn{1}{c}{moderate} & \multicolumn{1}{c}{hard}  & \multicolumn{1}{c}{easy} & \multicolumn{1}{c}{moderate} & \multicolumn{1}{c}{hard}  & \multicolumn{1}{c}{easy} & \multicolumn{1}{c}{moderate} & \multicolumn{1}{c}{hard} &                           \\ \midrule
                                   \ccross   &   \ccross                                     & 88.68                     & 78.85                     & \multicolumn{1}{c}{74.36} & 56.30                     & 49.13                     & \multicolumn{1}{c}{43.33} & 76.48                     & 58.62                     & 54.79                     & 64.50 \\
                                 \ccross   &   \ccheck                                     &       \textbf{88.98}              &    77.80                  &  \multicolumn{1}{c}{73.81} &   56.53                   &   49.73                   & \multicolumn{1}{c}{44.61} &   81.50                   &   61.74                   &        57.67             &  65.82\\
                                $\ccheck$      &   \ccross                                     & 87.12                     & 77.34                     & \multicolumn{1}{c}{74.63} & \textbf{58.66}                    & \textbf{50.34}                     & \textbf{45.19} & 81.09                    & 61.71                     & 58.00                     &66.01 \\
                                  $\ccheck$     &    $\ccheck$                                     & 88.79                    & \textbf{78.93}                    & \textbf{75.41} & 56.02                     & 48.55                      & 43.19 & \textbf{82.85}                     & \textbf{64.40}                     & \textbf{60.53}                     & \textbf{66.52} \\ \bottomrule
\end{tabular}} %
\caption{The ablation study of the spatial and temporal equivariance evaluated on the task of object detection with VoxelRCNN \cite{deng2021voxel}. The reported numbers are 3D mean average precision (\%) for the ``Car" ``Pedestrian" and ``Cyclist" categories for the 3 difficulty levels and 40 recall positions.}
\label{tab:ablation}
\end{table*}

\noindent\textbf{Quantitative results.}
We evaluate our pre-training framework for object detection on two networks SECOND \cite{second} and VoxelRCNN \cite{deng2021voxel}.
These detectors share a common sparse convolutional 3D feature extraction backbone and are initialized with the same model pre-trained on KITTI-360. We compare these fine-tuning results against PointContrast, STRL, and ALSO, as well as fine-tuning from a random weight initialization, denoted as ``No pre-training." Table \ref{tab:second} shows results with the detector SECOND.  We perform best or second best across most categories, and outperform the invariant counterpart STRL \cite{huang2021spatio}. In Table \ref{tab:voxelrcnn}, we demonstrate performance on the detector VoxelRCNN \cite{deng2021voxel}. We perform best or second best in most categories. Overall, we outperform both PointContrast and STRL in general, showing that joint spatio-temporal equivariance is a good self-supervision signal for 3D object detection. We perform on-par with the recent state-of-the-art method ALSO. We note that for SECOND, ALSO's pre-training strategy trains both the 3D feature backbone as well as the 2D convolutional layers of the detection network, leaving only the classification and regression box prediction layers to be randomly initialized. On the other hand, we train only the 3D backbone and leave the rest of the network to be randomly initialized. Additionally, our method converges much more quickly than ALSO, which is trained for 75 epochs, whereas our approach converges at around 10-20 epochs. Importantly, that these two approaches use different types of self-supervised learning techniques -- ALSO uses a generative strategy while \method~ uses cleverly designed loss functions for representation learning.

\begin{figure}
    \centering
    \includegraphics[width=.45\linewidth]{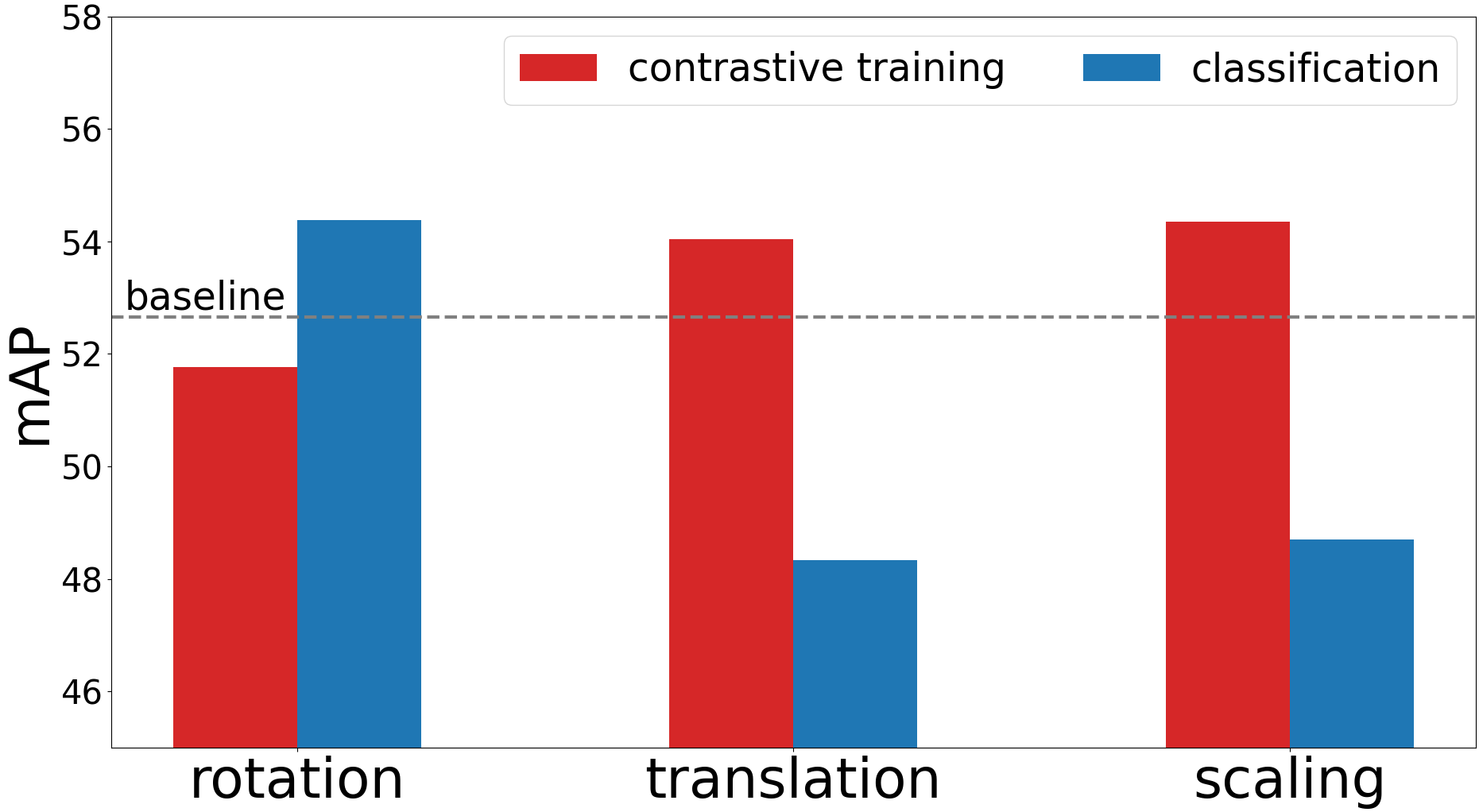}
    \caption{Relative 3D mean average precision of the object detector SECOND \cite{second} pre-trained for equivariance for the random spatial augmentations flip, rotation, translation, and scaling using the contrastive and classification objectives. The baseline network is pre-trained to be equivariant to the ``random flip" augmentation using contrastive learning.}
    \label{fig:prelim}
\end{figure}

\noindent\textbf{Ablation study.}
We conduct the ablation study on the spatial and temporal equivariance constraints and evaluate on the task of training VoxelRCNN \cite{deng2021voxel} on $5\%$ of KITTI data. We show the results in Table~\ref{tab:ablation} where we report the 3D mean average precision for all the three object categories for the three difficulty levels and 40 recall positions. By spatial equivariance, we mean training the network to be equivariant to the $n$-fold rotations using the cross-entropy loss and equivariant to random flips, scaling, and translations using the contrastive objective. By temporal equivariance, we mean training the network to be only be equivariant to 3D scene flow. Table~\ref{tab:ablation} shows that enforcing both the spatial and temporal equivariance constraints performs the best overall and that both equivariance constraints contribute to the performance. We observe that for the pedestrian category, pre-training with both objectives is not beneficial, but overall proves to be the better pre-training strategy.

\subsection{Choice of loss function for equivariant pre-training}
\label{sec:prelim}
We compare the effectiveness of the learning equivariance through contrastive training versus equivariance-via-classification as a pretext task for 3D object detection on KITTI with SECOND \cite{second}. We test the effect of replacing the contrastive objective for the classification objective for individual transformations in Fig.~\ref{fig:prelim}. Specifically, we use PointContrast \cite{xie2020pointcontrast} with a single ``random flip" augmentation as the baseline contrastive pretext task.  The baseline performance under PointContrast is indicated by the gray dotted line. We encourage the network to be equivariant to three types of rigid transformations using the two learning objectives. For each additional transformation, we train the network to either predict the transformation in addition to the baseline or only train the network under PointInfoNCE loss. In the case of random rotation and scaling, prediction is a 10-class classification problem. In the case of translation, we predict the translation along each axes using a multi-label multi-class loss objective. We observe that encouraging equivariance to global translation along each axis as well as to scaling purely through the contrastive objective improves the performance. On the other hand, training for equivariance to rotation using the classification objective boosts performance relative to using the PointInfoNCE loss. The right choice of loss function depends on the nature of the augmentation. The standard ranges for translation augmentation for LiDAR object detectors is ($0m,0.2m$) along each axis. Considering that the range of the KITTI dataset reaches $70m$, this is a difficult fine-grained prediction task. The range of the scaling transformation is similarly small, ($0.95,1.05$). For $n$-fold rotations, the scene is rotated along the vertical axis by an angle ranging from $(\frac{-\pi}{2},\frac{\pi}{2})$, a much larger range that results in more distinct augmentations. These experiments show that training the network to predict $n$-fold rotations while training the network under the point-level contrastive loss for the scaling, flip, and translation augmentations is a good strategy.

\subsection{Limitations}
We acknowledge certain limitations in our proposed self supervised learning framework. We observe that when fine-tuning on the KITTI dataset, self supervised pre-training does not always boost performance for the ``Car" category. We believe this is due to the fact that this category is well represented in the dataset, and is a relatively ``easier" object to detect. Additionally, \method~ does not always outperform the SOTA approach ALSO , and performs second-best for certain categories. However, we point out that ALSO is a generative approach, which is not directly comparable to our method, which also may be integrated with our discriminative method, which we hope to explore in the future. We also observe that when plenty of annotated samples are available (\eg, 100\% of data in Tables \ref{tab:second} and \ref{tab:voxelrcnn}), pre-training does not have a large impact on performance, and we emphasize that \method~ is most helpful in low-data scenarios, and it achieves close to full-data performance with just $20\%$ of annotated training data in the case of KITTI object detection dataset.

\section{Conclusion}
In this work, we examine the role of equivariance in representation learning for large scale outdoor point clouds. We present \method~, a self supervised learning method for 3D object detection on LiDAR scenes that learns meaningful geometric representation by encouraging joint spatial and temporal equivariance. We developed a novel 3D scene flow equivariance objective to incorporate temporal information for improved representation learning. We showed that the choice of equivariance objective affects the final performance significantly depending on the type of augmentation applied. Our experiments demonstrate the usefulness of our learned representations and suggest that for certain transformations, it is helpful to encourage equivariance through augmentation classification.

{
    \small
    \bibliographystyle{splncs04}
    \bibliography{main}

\begin{thebibliography}{10}
\providecommand{\url}[1]{\texttt{#1}}
\providecommand{\urlprefix}{URL }
\providecommand{\doi}[1]{https://doi.org/#1}

\bibitem{behley2019iccv}
Behley, J., Garbade, M., Milioto, A., Quenzel, J., Behnke, S., Stachniss, C., Gall, J.: {SemanticKITTI: A Dataset for Semantic Scene Understanding of LiDAR Sequences}. In: Proc. of the IEEE/CVF International Conf.~on Computer Vision (ICCV) (2019)

\bibitem{bhardwaj2023steerable}
Bhardwaj, S., McClinton, W., Wang, T., Lajoie, G., Sun, C., Isola, P., Krishnan, D.: Steerable equivariant representation learning. arXiv preprint arXiv:2302.11349  (2023)

\bibitem{boulch2023also}
Boulch, A., Sautier, C., Michele, B., Puy, G., Marlet, R.: {ALSO}: Automotive lidar self-supervision by occupancy estimation. In: Proceedings of the IEEE/CVF Conference on Computer Vision and Pattern Recognition. pp. 13455--13465 (2023)

\bibitem{chen2020simple}
Chen, T., Kornblith, S., Norouzi, M., Hinton, G.: A simple framework for contrastive learning of visual representations. In: International conference on machine learning. pp. 1597--1607. PMLR (2020)

\bibitem{choy20194d}
Choy, C., Gwak, J., Savarese, S.: {4D} spatio-temporal {ConvNets}: Minkowski convolutional neural networks. In: Proceedings of the IEEE/CVF conference on computer vision and pattern recognition. pp. 3075--3084 (2019)

\bibitem{dangovski2021equivariant}
Dangovski, R., Jing, L., Loh, C., Han, S., Srivastava, A., Cheung, B., Agrawal, P., Solja{\v{c}}i{\'c}, M.: Equivariant contrastive learning. arXiv preprint arXiv:2111.00899  (2021)

\bibitem{deng2021voxel}
Deng, J., Shi, S., Li, P., Zhou, W., Zhang, Y., Li, H.: {Voxel R-CNN}: Towards high performance voxel-based {3D} object detection. In: Proceedings of the AAAI Conference on Artificial Intelligence. vol.~35, pp. 1201--1209 (2021)

\bibitem{devillers2022equimod}
Devillers, A., Lefort, M.: {EquiMod}: An equivariance module to improve self-supervised learning. arXiv preprint arXiv:2211.01244  (2022)

\bibitem{fan2022embracing}
Fan, L., Pang, Z., Zhang, T., Wang, Y.X., Zhao, H., Wang, F., Wang, N., Zhang, Z.: Embracing single stride {3D} object detector with sparse transformer. In: Proceedings of the IEEE/CVF conference on computer vision and pattern recognition. pp. 8458--8468 (2022)

\bibitem{garrido2023self}
Garrido, Q., Najman, L., Lecun, Y.: Self-supervised learning of split invariant equivariant representations. arXiv preprint arXiv:2302.10283  (2023)

\bibitem{KITTI}
Geiger, A., Lenz, P., Urtasun, R.: Are we ready for autonomous driving? {T}he {KITTI} vision benchmark suite. In: Conference on Computer Vision and Pattern Recognition (CVPR) (2012)

\bibitem{grill2020bootstrap}
Grill, J.B., Strub, F., Altch{\'e}, F., Tallec, C., Richemond, P., Buchatskaya, E., Doersch, C., Avila~Pires, B., Guo, Z., Gheshlaghi~Azar, M., et~al.: Bootstrap your own latent: A new approach to self-supervised learning. Advances in neural information processing systems  \textbf{33},  21271--21284 (2020)

\bibitem{gupta2023learning}
Gupta, S., Robinson, J., Lim, D., Villar, S., Jegelka, S.: Learning structured representations with equivariant contrastive learning  (2023)

\bibitem{he2021masked}
He, K., Chen, X., Xie, S., Li, Y., Doll’ar, P., Girshick, R.B.: Masked autoencoders are scalable vision learners. 2022 ieee. In: CVF Conference on Computer Vision and Pattern Recognition (CVPR). pp. 15979--15988 (2021)

\bibitem{he2020momentum}
He, K., Fan, H., Wu, Y., Xie, S., Girshick, R.: Momentum contrast for unsupervised visual representation learning. In: Proceedings of the IEEE/CVF conference on computer vision and pattern recognition. pp. 9729--9738 (2020)

\bibitem{huang2021spatio}
Huang, S., Xie, Y., Zhu, S.C., Zhu, Y.: Spatio-temporal self-supervised representation learning for 3d point clouds. In: Proceedings of the IEEE/CVF International Conference on Computer Vision. pp. 6535--6545 (2021)

\bibitem{jin2022deformation}
Jin, Z., Lei, Y., Akhtar, N., Li, H., Hayat, M.: Deformation and correspondence aware unsupervised synthetic-to-real scene flow estimation for point clouds. In: Proceedings of the IEEE/CVF Conference on Computer Vision and Pattern Recognition. pp. 7233--7243 (2022)

\bibitem{lang2019pointpillars}
Lang, A.H., Vora, S., Caesar, H., Zhou, L., Yang, J., Beijbom, O.: {PointPillars}: Fast encoders for object detection from point clouds. In: Proceedings of the IEEE Conference on Computer Vision and Pattern Recognition. pp. 12697--12705 (2019)

\bibitem{Liao2022PAMI}
Liao, Y., Xie, J., Geiger, A.: {KITTI}-360: A novel dataset and benchmarks for urban scene understanding in 2d and 3d. Pattern Analysis and Machine Intelligence (PAMI)  (2022)

\bibitem{mao2021one}
Mao, J., Niu, M., Jiang, C., Liang, H., Chen, J., Liang, X., Li, Y., Ye, C., Zhang, W., Li, Z., et~al.: One million scenes for autonomous driving: {ONCE} dataset. arXiv preprint arXiv:2106.11037  (2021)

\bibitem{nunes2022segcontrast}
Nunes, L., Marcuzzi, R., Chen, X., Behley, J., Stachniss, C.: {SegContrast}: {3D} point cloud feature representation learning through self-supervised segment discrimination. IEEE Robotics and Automation Letters  \textbf{7}(2),  2116--2123 (2022)

\bibitem{nunes2023temporal}
Nunes, L., Wiesmann, L., Marcuzzi, R., Chen, X., Behley, J., Stachniss, C.: Temporal consistent {3D} {LiDAR} representation learning for semantic perception in autonomous driving. In: Proceedings of the IEEE/CVF Conference on Computer Vision and Pattern Recognition. pp. 5217--5228 (2023)

\bibitem{shi2020pv}
Shi, S., Guo, C., Jiang, L., Wang, Z., Shi, J., Wang, X., Li, H.: {PV-RCNN}: Point-voxel feature set abstraction for {3D} object detection. In: Proceedings of the IEEE/CVF Conference on Computer Vision and Pattern Recognition. pp. 10529--10538 (2020)

\bibitem{shi2023pv}
Shi, S., Jiang, L., Deng, J., Wang, Z., Guo, C., Shi, J., Wang, X., Li, H.: {PV-RCNN++}: Point-voxel feature set abstraction with local vector representation for {3D} object detection. International Journal of Computer Vision  \textbf{131}(2),  531--551 (2023)

\bibitem{shi2019pointrcnn}
Shi, S., Wang, X., Li, H.: {PointRCNN: 3D} object proposal generation and detection from point cloud. In: Proceedings of the IEEE Conference on Computer Vision and Pattern Recognition. pp. 770--779 (2019)

\bibitem{shi2021parta2}
Shi, S., Wang, Z., Shi, J., Wang, X., Li, H.: From points to parts: {3D} object detection from point cloud with part-aware and part-aggregation network. IEEE Transactions on Pattern Analysis and Machine Intelligence  \textbf{43}(08),  2647--2664 (2021)

\bibitem{waymo}
Sun, P., Kretzschmar, H., Dotiwalla, X., Chouard, A., Patnaik, V., Tsui, P., Guo, J., Zhou, Y., Chai, Y., Caine, B., et~al.: Scalability in perception for autonomous driving: Waymo open dataset. In: Proceedings of the IEEE/CVF Conference on Computer Vision and Pattern Recognition. pp. 2446--2454 (2020)

\bibitem{tang2020searching}
Tang, H., Liu, Z., Zhao, S., Lin, Y., Lin, J., Wang, H., Han, S.: Searching efficient {3D} architectures with sparse point-voxel convolution. In: European conference on computer vision. pp. 685--702. Springer (2020)

\bibitem{openpcdet2020}
Team, O.D.: {OpenPCDet}: An open-source toolbox for {3D} object detection from point clouds. \url{https://github.com/open-mmlab/OpenPCDet} (2020)

\bibitem{wei2021pv}
Wei, Y., Wang, Z., Rao, Y., Lu, J., Zhou, J.: {PV-RAFT}: Point-voxel correlation fields for scene flow estimation of point clouds. In: Proceedings of the IEEE/CVF conference on computer vision and pattern recognition. pp. 6954--6963 (2021)

\bibitem{xiao2020should}
Xiao, T., Wang, X., Efros, A.A., Darrell, T.: What should not be contrastive in contrastive learning. arXiv preprint arXiv:2008.05659  (2020)

\bibitem{xie2020self}
Xie, Q., Luong, M.T., Hovy, E., Le, Q.V.: Self-training with noisy student improves {ImageNet} classification. In: Proceedings of the IEEE/CVF Conference on Computer Vision and Pattern Recognition. pp. 10687--10698 (2020)

\bibitem{xie2020pointcontrast}
Xie, S., Gu, J., Guo, D., Qi, C.R., Guibas, L., Litany, O.: {PointContrast}: Unsupervised pre-training for {3D} point cloud understanding. In: Computer Vision--ECCV 2020: 16th European Conference, Glasgow, UK, August 23--28, 2020, Proceedings, Part III 16. pp. 574--591. Springer (2020)

\bibitem{xiong2021self}
Xiong, Y., Ren, M., Zeng, W., Urtasun, R.: Self-supervised representation learning from flow equivariance. In: Proceedings of the IEEE/CVF International Conference on Computer Vision. pp. 10191--10200 (2021)

\bibitem{xu2023mv}
Xu, R., Wang, T., Zhang, W., Chen, R., Cao, J., Pang, J., Lin, D.: {MV-JAR}: Masked voxel jigsaw and reconstruction for {LiDAR}-based self-supervised pre-training. In: Proceedings of the IEEE/CVF Conference on Computer Vision and Pattern Recognition. pp. 13445--13454 (2023)

\bibitem{yan2023spot}
Yan, X., Chen, R., Zhang, B., Yuan, J., Cai, X., Shi, B., Shao, W., Yan, J., Luo, P., Qiao, Y.: {SPOT}: Scalable {3D} pre-training via occupancy prediction for autonomous driving. arXiv preprint arXiv:2309.10527  (2023)

\bibitem{second}
Yan, Y., Mao, Y., Li, B.: {SECOND}: Sparsely embedded convolutional detection. Sensors  \textbf{18}(10) (2018). \doi{10.3390/s18103337}, \url{https://www.mdpi.com/1424-8220/18/10/3337}

\bibitem{yang2023gd}
Yang, H., He, T., Liu, J., Chen, H., Wu, B., Lin, B., He, X., Ouyang, W.: {GD-MAE}: generative decoder for {MAE} pre-training on {LiDAR} point clouds. In: Proceedings of the IEEE/CVF Conference on Computer Vision and Pattern Recognition. pp. 9403--9414 (2023)

\bibitem{yang20203dssd}
Yang, Z., Sun, Y., Liu, S., Jia, J.: {3DSSD}: Point-based {3D} single stage object detector. In: Proceedings of the IEEE/CVF Conference on Computer Vision and Pattern Recognition. pp. 11040--11048 (2020)

\bibitem{yin2022proposalcontrast}
Yin, J., Zhou, D., Zhang, L., Fang, J., Xu, C.Z., Shen, J., Wang, W.: {ProposalContrast}: Unsupervised pre-training for {LiDAR}-based {3D} object detection. In: European Conference on Computer Vision. pp. 17--33. Springer (2022)

\bibitem{zhang2021self}
Zhang, Z., Girdhar, R., Joulin, A., Misra, I.: Self-supervised pretraining of {3D} features on any point-cloud. In: Proceedings of the IEEE/CVF International Conference on Computer Vision. pp. 10252--10263 (2021)

\bibitem{zhu2021cylindrical}
Zhu, X., Zhou, H., Wang, T., Hong, F., Ma, Y., Li, W., Li, H., Lin, D.: Cylindrical and asymmetrical {3D} convolution networks for {LiDAR} segmentation. In: Proceedings of the IEEE/CVF conference on computer vision and pattern recognition. pp. 9939--9948 (2021)

\end{thebibliography}
}

\newpage

\section{Supplementary Material}
We present supplementary material for this paper. We present additional quantitative results of \method~ when pre-trained on the Waymo Open Dataset \cite{waymo} in Section \ref{sec:waymo}.

\subsection{Pre-training on the Waymo Open Dataset}
\label{sec:waymo}
We present additional experimental results of training \method~ on the Waymo Open Dataset (WOD) \cite{waymo} as a pre-training step and fine-tuning on the KITTI object detection dataset \cite{KITTI}. As in the main paper, we perform fine-tuning on two popular object detection networks SECOND \cite{second} and VoxelRCNN \cite{deng2021voxel} on 3 different splits of data, $\{5\%,20\%,100\%\}$. The pre-training dataset consists of $100k$ samples from the WOD cropped to a front field-of-view to be consistent with the main experimental settings. In order to mitigate the distribution gap between datasets, we train the networks on $\{x,y,z\}$ coordinates, leaving out intensity. We compare the performance of our approach with that of PointContrast \cite{xie2020pointcontrast} and STRL \cite{huang2021spatio}. As the weights of ALSO \cite{boulch2023also} trained on WOD are unavailable, we forgo this comparison.

In Table \ref{tab:waymo_second}, we present the results of fine-tuning SECOND on three data splits. We consistently perform best or second best across all data splits. Particularly in the $5\%$ and $100\%$ data splits, we out-perform comparative methods by a large margin. In Table \ref{tab:waymo_voxelrcnn}, we present the results of fine-tuning VoxelRCNN on three data splits. We consistently perform best or second best across almost all data splits. We observe that when training with $100\%$ of data, none of the pre-training strategies are beneficial for VoxelRCNN, however, we see significant improvements in the other data splits, reinforcing the utility of our framework as a pre-training strategy in low-data scenarios.

\begin{table*}

\centering
\resizebox{\textwidth}{!}{
\begin{tabular}{cccccccccccc}
\toprule 
\multirow{3.5}{*}{Split} & \multirow{3.8}{*}{Method} & \multicolumn{9}{c}{average precision (AP) (\%)}                                             & \multirow{3.8}{*}{mAP (\%)} \\ \cmidrule{3-11}
                       &                                              & \multicolumn{3}{c}{Car} & \multicolumn{3}{c}{Pedestrian} & \multicolumn{3}{c}{Cyclist} &                      \\ \cmidrule(l){3-5} \cmidrule(l){6-8} \cmidrule(l){9-11}
                       &                                              & easy   & moderate   & hard  & easy     & moderate     & hard     & easy     & moderate    & hard    &                      \\ \midrule
\multirow{4}{*}{5\%}   & No pre-training                              & \bf{86.40}  & 73.01  & 67.79 & 42.76    & \ul{38.26}    & 34.42    & 64.13    & 45.44   & 42.70   & 54.99                \\
                       & PointContrast                            &  \ul{86.32} & \ul{73.14}	 & \ul{69.73} & \ul{42.78}  &  36.93	  &  33.47	   &  \ul{70.51}	   &  49.76	  &  46.77	  &  56.60                  \\
                       & STRL                                     & 86.23  &  72.19	 & 68.67	  & 43.13	    &  38.07	   & \ul{34.62}	   &  70.43	   & \ul{50.80}	  & 47.63	   &  \ul{56.86}               \\
                       
                       & \method~                                  & 86.30	 & \bf{73.34}	  &  \bf{69.91}	&  \bf{45.36}	   & \bf{39.57}	    & \bf{35.86}	   & \bf{72.44}	    & \bf{51.51}    & \bf{48.39}   &   \bf{58.07}            \\ \midrule
\multirow{4}{*}{20\%}  & No pre-training                              & 87.64  & 77.12  & 73.13 & 50.36    & 45.70    & 41.29    & 75.26    & 54.99   & 51.24   & 61.86                \\
                       & PointContrast                            & \bf{90.42}	 & \bf{79.09}	 & \bf{76.13}	 &  \bf{52.18}	  & \bf{46.68}	    & \bf{42.91}	  &  \bf{80.87}	  & \bf{63.75}	   & \bf{59.77}	    &   \bf{65.75}              \\
                       & STRL                                     &  89.64 & \ul{78.44} & 75.53	  &  50.64	   &  45.31	  &  41.58	   & 79.22	    & 61.59	  &  57.92	  &     64.43         \\
                       
                       & \method~                                 &   \ul{89.71}  & \ul{78.44}	 &  \ul{75.67}	& \ul{50.83}	  &   \ul{45.77}	  & \ul{41.94}	  & \ul{80.29}	   & \ul{62.64}	  & \ul{58.91}	    & \ul{64.91} \\ \midrule 
\multirow{4}{*}{100\%} & No pre-training                              & \ul{90.05}  & \ul{81.03}  & \ul{78.09} & \ul{53.97}    & \ul{49.19}   & \ul{44.06}    & \ul{80.59}    & \ul{63.54}   & \ul{59.64}   & \ul{66.69}                \\
                       & PointContrast                            & 88.06	 & 78.98	 & 76.05	 & 52.85	   & 47.18	     & 42.29	  & 79.04	   & 60.80	   &  56.85	  &   64.68              \\
                       & STRL                                     & 90.04	  & 80.83	 &  77.66	&  53.44	   &  48.39	  &  43.13	   & 78.87	    & 59.88	 & 55.87	   &     65.35         \\
                       
                       & \method~                                  &  \bf{90.38}      & \bf{81.31}	       & \bf{78.19}	      &   \bf{55.76}     &  \bf{50.61}	       & \bf{46.08}	          &   \bf{81.49}	    &  \bf{63.59}	      &   \bf{59.69}	   &  \bf{67.46}                   \\  \bottomrule
\end{tabular} %
} 
\caption{3D object detection with SECOND \cite{second} pre-trained on the Waymo Open Dataset \cite{waymo} and fine-tuned on KITTI \cite{KITTI} under different data splits. Each result is an average over 3 fixed subsets of the dataset. We report 3D average precision for 3 categories as well as the mean average precision over 40 recall positions. The best and second best performance is marked in \bd{bold} and \ul{underline}, respectively.}
\label{tab:waymo_second}
\end{table*}

\begin{table*}
\centering
\resizebox{\textwidth}{!}{
\begin{tabular}{ccccccccccccc}
\toprule  %
\multirow{3.5}{*}{Split} & \multirow{3.8}{*}{Method}  & \multicolumn{9}{c}{average precision (AP) (\%)}                                             & \multirow{3.8}{*}{mAP (\%)} \\ \cmidrule{3-11}
                       &                                             & \multicolumn{3}{c}{Car} & \multicolumn{3}{c}{Pedestrian} & \multicolumn{3}{c}{Cyclist} &                      \\ \cmidrule(l){3-5} \cmidrule(l){6-8} \cmidrule(l){9-11}
                       &                                               & easy   & moderate   & hard  & easy     & moderate     & hard     & easy    & moderate    & hard    &                      \\ \midrule
\multirow{4}{*}{5\%}   & No pre-training                              & 88.89  & \bf{79.21}  & \bf{75.55} & \bf{57.50}    & \bf{49.84}    & \bf{44.27}    & 78.92   & 59.73   & 55.97   & 65.54                \\
                       & PointContrast                            & 88.25 & 76.30	&71.65	 & 51.90	   &  44.37	 &  40.01	   & 80.67	   &   60.60	 & 56.54	   &   63.37             \\
                       & STRL                                     & \bf{89.15}	   &  77.29	 & 73.73	 &  56.04	   &  \ul{49.13}	   & 43.59	    &  83.55	  & \bf{63.81}	   & \bf{59.61}	    & \bf{66.21}                \\
                       
                       & \method~                                  &   \ul{89.13}	     & \ul{77.33}	     &  \ul{73.84}	     &   \bf{56.06}	    &  48.87	        &  \ul{43.70}	       & \bf{83.57}	        &  \ul{63.28}	      &  \ul{59.12}	       &   \ul{66.10}                    \\  \midrule
\multirow{4}{*}{20\%}  & No pre-training                              & \bf{91.99}  & \bf{82.10}  & \bf{79.40} & 56.09    & 49.29    & 44.26    & 85.24   & \bf{67.55}   & \bf{63.13}   & \ul{68.78}                \\
                       & PointContrast                            & 91.74	  &  80.47	 & 77.35	  &  \bf{59.30}	   &  \ul{51.05}	  & \ul{45.90}	    &  \ul{85.97}	  & 65.70	   & 61.25	   &   68.75              \\
                       & STRL                                     & \ul{91.95}	  &  \ul{81.04}	 & \ul{77.89}	 &  58.25	   &   50.53	  & 45.37	    &  85.36	 &  \ul{66.24}	  & 62.00	   &   68.74              \\
                      
                       & \method~                                  & 91.74	  & 80.46	  & 77.27	 & \ul{59.26}	    &   \bf{51.82}   &  \textbf{46.65}	  &\bf{86.51}   & 67.44	 &  \ul{62.86}	  &   \bf{69.33}              \\                 \midrule %
\multirow{4}{*}{100\%} & No pre-training                              & \bf{92.45}  & \bf{83.00}  & \bf{80.20} & \bf{62.41}    & \bf{55.89}    & \bf{50.31}    & 88.40   & 68.81   & 64.42   & \bf{71.77}                \\
                       & PointContrast                            & 91.61  & 82.26	 & 79.76	 & 55.47	  & 48.06	    &  43.28	  & \bf{89.68}	 & \bf{71.90}	   &  \bf{67.57}	  &     69.95          \\
                       & STRL                                     & 91.86	  & \ul{82.29} & \ul{79.80}	 &  \ul{59.65}	   & \ul{51.82}	   & \ul{46.23}	     & 87.28	  & 70.49	  &  65.79	  &    \ul{70.58}            \\
                       
                       & \method~                                  &   \ul{92.16}     &   82.16	     &  79.77	    &  59.14	        &    50.45	      &  45.04	        & \ul{88.68}     &    \ul{71.17}	    &  \ul{66.44}	      &    70.56                 \\  \bottomrule
\end{tabular}
} 

\caption{3D object detection with VoxelRCNN \cite{deng2021voxel} pre-trained on the Waymo Open Dataset \cite{waymo} and fine-tuned on KITTI \cite{KITTI} under different data splits. Each result is an average over 3 fixed subsets of the dataset. We report 3D average precision for 3 categories as well as the mean average precision over 40 recall positions. The best and second best performance is marked in \bd{bold} and \ul{underline}, respectively. }
\label{tab:waymo_voxelrcnn}
\end{table*}

\end{document}